\documentclass[sn-mathphys,Numbered]{sn-jnl}

\usepackage{graphicx}%
\usepackage{multirow}%
\usepackage{amsmath,amssymb,amsfonts}%
\usepackage{amsthm}%
\usepackage{mathrsfs}%
\usepackage[title]{appendix}%
\usepackage{xcolor}%
\usepackage{textcomp}%
\usepackage{manyfoot}%
\usepackage{booktabs}%
\usepackage{algorithm}%
\usepackage{algorithmicx}%
\usepackage{algpseudocode}%
\usepackage{listings}%
\usepackage{graphicx}
\usepackage{tikz}
\usepackage{lingmacros}
\usepackage{tree-dvips}
\usepackage{pgfplots}
\usepackage{xlop}
\pgfplotsset{width=7cm,compat=1.9}
\usepackage{scalerel}

\theoremstyle{thmstyleone}%
\theoremstyle{thmstyletwo}%

\theoremstyle{thmstylethree}%

\begin{document}

\title[Article Title]{``Turing Tests'' For An AI Scientist}


\author*[1]{\fnm{Xiaoxin} \sur{Yin}}


\abstract{The rapid advancements in deep learning have demonstrated the potential for AI agents to perform tasks previously limited to humans, including scientific research. While LLMs have shown impressive capabilities in solving math or coding problems, the ability to make scientific discoveries remains a distinct challenge. This paper proposes a "Turing test for an AI scientist" to assess whether an AI agent can conduct scientific research independently, without relying on human-generated knowledge.

Drawing inspiration from the historical development of science, we propose seven benchmark tests that evaluate an AI agent's ability to make groundbreaking discoveries in various scientific domains. These tests include inferring the heliocentric model from celestial observations, discovering the laws of motion in a simulated environment, deriving the differential equation governing vibrating strings, inferring Maxwell's equations from electrodynamics simulations, inventing numerical methods for initial value problems, discovering Huffman coding for data compression, and developing efficient sorting algorithms.

To ensure the validity of these tests, the AI agent is provided with interactive libraries or datasets specific to each problem, without access to human knowledge that could potentially contain information about the target discoveries. The ultimate goal is to create an AI scientist capable of making novel and impactful scientific discoveries, surpassing the best human experts in their respective fields. These "Turing tests" serve as intermediate milestones, assessing the AI agent's ability to make discoveries that were groundbreaking in their time.

If an AI agent can pass the majority of these seven tests, it would indicate significant progress towards building an AI scientist, paving the way for future advancements in autonomous scientific discovery. This paper aims to establish a benchmark for the capabilities of AI in scientific research and to stimulate further research in this exciting field.}




\maketitle

\section{Introduction}\label{intro}

The recent advances in deep learning, especially those in large language models, have shown the possibility of an AI agent performing any task a human can perform, including scientific research. Recent studies have shown that LLMs such as GPT-4\cite{gpt-4}, Microsoft Copilot\cite{ms-copilot}, and CodeLlama\cite{roziere23} can solve competition-level coding problems \cite{huang23}, and LLMs such as GPT-4 and Llemma\cite{azerbayev23} can solve some high-school-level competition math problems (including some IMO-level problems). These LLMs can certainly help researchers solve some problems they encounter in their daily research. 

However, being able to solve a type of well-defined problems is very different from making discoveries in scientific research. For instance, in order to train an LLM to solve coding problems, a general-purpose LLM is often fine-tuned on all public code on GitHub, and also fine-tuned on hundreds of thousands of coding problems from various platforms such as CodeForce and LeetCode. For example, CodeLlama-Python  underwent fine-tuning with 100 billion tokens of Python code. The LLM simply learns how to write code given the coding problem (which is the prompt), by learning to predict the next token in its code given the prompt and tokens it has generated. This is essentially the same methodology used to train a model to write novels after reading millions of novels. It does not have the capability of discovering what it has not been taught, making it unable to make scientific discoveries like a scientist would do.

This makes it necessary to define a ``qualification test for an AI scientist''. If an AI agent can finish this test without help from human, we can conclude that this agent qualifies as a scientist and can conduct scientific research on its own. 

This resembles the Turing Test, which was proposed by Alan Turing in 1950 and serves as a foundational concept in the field of artificial intelligence, challenging whether machines can exhibit human-like intelligence. Turing's seminal paper, ``Computing Machinery and Intelligence"\cite{Turing1950}, introduced the idea of an imitation game where a human interrogator would attempt to distinguish between a computer and a human through a series of text-based questions. The inability of the interrogator to consistently identify the machine is considered a measure of the machine's intelligence. This test not only sparked decades of philosophical debate but also drove technological advances in AI research, shaping the development of intelligent systems.

Unlike today's LLMs which are trained on a very large corpus in order to perform similar tasks, science is about discoveries, especially in new areas that have not been explored. In order to define a Turing test for an AI scientist, let us first review the development of science in its early stage. 

The night sky played an essential role in the transition to modern scientific methodologies, largely through the efforts of astronomers such as Johannes Kepler and Galileo Galilei. Kepler's laws of planetary motion, derived from meticulous observations of the night sky, laid the groundwork for the heliocentric model of the solar system and ultimately for Newton's theory of gravitation. His reliance on empirical data and systematic experimentation marked a significant departure from the speculative philosophies that had previously dominated the scientific arena. Galileo's method of integrating experimental evidence with mathematical analysis is a cornerstone of the scientific method, earning him the title ``father of modern science." His work exemplifies how observations of the night sky were instrumental in shaping the development of science in its modern form.

Therefore, the first ``Turing test'' for an AI scientist should be the discovery of the heliocentric model through the observations of the night sky. This requires an AI agent to discover laws governing the motions of celestial objects, and fit them into a mathematical framework. It also requires the AI agent to make groundbreaking conjectures such as the earth is similar to the planets in the night sky. Both requirements are necessities for a scientist.

In order to be a good benchmark test for an AI scientist, a test needs to provide a very large amount of data or an interactive environment. For example, one can access the location of any observable celestial object at any moment of time through the AstroPy library\cite{astropy2022}. 

Based on the above two standards we choose the following seven tests as the Turing tests for an AI scientist. In each test the AI agent cannot be trained on human knowledge, but is accessible to math tools such as SymPy\cite{sympy} and NumPy\cite{numpy}, and any other datasets that do not ``leak information'', i.e., containing clues of target discoveries to be made.

\begin{enumerate}
	\item \emph{Heliocentric Model}: Given an interactive python library\cite{astropy2022} that provides the coordinates of any observable celestial object in the night sky at any given moment, check if an AI agent can infer Kepler's three laws and conclude that all planets orbit the sun. A bonus question is that the earth orbits the sun but it is not required. 
	
	\item \emph{Laws of Motions}: Given an interactive library that controls Minecraft\cite{mcpi}, check if an AI agent can discover the Law of Inertia and the Law of Acceleration (only for gravity).
	
	\item \emph{Vibrating Strings}: Vibrating strings is one of the most important problems that drove the development of differential equations\cite{kurrer2012history}. Given a Python library that provides the position of each point on a vibrating string of many different initial conditions, check if an AI agent can infer the differential equation governing the motion:
	\begin{equation*}
		\frac{\partial^2 u}{\partial t^2} = c^2 \frac{\partial^2 u}{\partial x^2}
	\end{equation*}
	where \( u(x,t) \) is the displacement of the string, \( c \) is the speed of wave propagation in the string, \( t \) is time, and \( x \) is the spatial coordinate along the string. Please note the AI agent should not have any prior knowledge about calculus, and has to define differential equations on its own.
	
	\item \emph{Maxwell's Equations}: Maxwell's equations are often considered to be the most beautiful equations in physics. Given a Python-based electrodynamics simulator\cite{fdtd}, check if an AI agent can infer the Maxwell's equations or their equivalent forms. Again the agent cannot use any prior knowledge about calculus.
	
	\item \emph{Initial Value Problem (IVP)}: IVP is probably the most important problem in numerical computing, and the Runge-Kutta method\cite{Lambert1991} invented at the end of the 19th century is still widely used today. Given math tools such as SymPy\cite{sympy} and NumPy\cite{numpy} that can calculate integrals of functions both symbolically and numerically, check if an AI agent can invent a method for IVP that is at least as accurate as the fourth-order Runge-Kutta method. 
	
	\item \emph{Huffman Coding}: Huffman coding\cite{Huffman1952} is a most important piece of work in information theory. Given a large corpus of ascii characters, and Python functions to operate on bits, check if an AI agent can discover Huffman coding when working towards the goal of minimizing storage under the constraint that each character be represented by a specific sequence of 0's and 1's.
	
	\item \emph{Sorting Algorithm}: Sorting is probably the most studied problem in computer science. Given a very large number of examples of sorting integer arrays and a Python environment, check if an AI can discover a sorting algorithm that runs in expected $O(n \log{n})$ time.
	
\end{enumerate}

Please note that each test selected only requires data or interaction within a well-defined scope (such as a dataset or an interactive library). This makes it possible for an AI agent to make discoveries without being trained on human-written documents, which may leak information about the target discoveries. For the same reason we do not select any tests from many most important disciplines, such as chemistry, biology, and geology, because they either require interacting with the physical world or have a limited amount of observations. In order to make important discoveries in these disciplines, it is inevitable to use knowledge outside a small predefined scope, which may leak key information to the AI agent.

The ultimate goal for an AI scientist should be making novel and impactful scientific discoveries that no one has made before. Then why do we still need these ``Turing tests'' which have been discovered decades or centuries ago? The reason is that the ``ultimate goal'' is very challenging because the AI agent needs to be better than the best human experts in the world. It is analogical to building an AI agent that can beat the best GO player in the world, while our benchmark is like beating a top GO player a thousand years ago when GO was in its early age, or beating an amateur GO player today. If we could build an AI agent that passes the majority of the above seven tests, we can conclude that we are in the right direction of building an AI scientist, and it should evolve into someone who can make important scientific discoveries in the foreseeable future.

\section{Related Work}\label{relwork}

The idea of automating scientific research activities dates back to the early days of computer science. An article on \emph{Science} in 2009 \cite{waltz09} provides a great overview on the early explorations. Also in 2009 a ``Robot Scientist'' named Adam was released \cite{king09}. The authors developed specialized hardware for conducting basic experiments, such as tracking yeast growth with varying gene deletions and metabolites. This was paired with logic programming software for selecting experiments. The software keeps track of various hypotheses and chooses experiments likely to refute many of them at once. These experiments are automatically performed, and their results guide the next experiment's selection. Adam effectively identified the functions of multiple genes, requiring fewer experiments compared to other experiment-selection methods like cost-based choices. \cite{naik16} presents a research that utilizes special hardwares to automatically learn the effects of different drugs upon the distribution of different proteins within mammalian cells.

Very recently a breakthrough was brought by DeepMind \cite{trinh24}, in which the authors created a large language model that learned geometry on one billion generated problems, in order to discover geometry properties, and train itself to prove these properties. The model was tested on 30 IMO geometry and got 25 of them correct, which outperforms the majority of IMO participants. This is the first time a neural network model learns to master a discipline of science on its own, and it will not be surprising if the same methodology can be extended to other disciplines such as number theory and combinatorics. 

Our goal is to let AI make scientific discoveries on its own. There are two routes towards this goal. The first is to build an AI agent that can make novel and impactful scientific discoveries that have not been made before. This is our ultimate goal. But it is very challenging because the AI agent needs to be better than the best human expert in a field. 

The alternative route is to build an AI agent that can make some of the most important scientific discoveries in the history, without reading human knowledge that may contain key information to these discoveries. We believe this is an easier route because some of such discoveries can be inferred from abundant data and a scientific methodology. Comparing with the first goal which is analogical to building an AI agent that can beat the best GO player in the world, the second goal is like building an AI agent that can beat an amateur GO player. We believe the second goal is a good starting point for building an AI scientist, which should eventually evolve into someone who can make new and important scientific discoveries.

\section{The Seven Qualification Tests for an AI Scientist}\label{tests}

\subsection{Selection Criteria}

An ideal ``Turing'' test for an AI scientist should satisfy the following three criteria:

\begin{enumerate}
	\item It is the key to an important discovery in the development of science.
	\item It is possible to be discovered digitally, without interaction with the physical world. 
	\item The discovery is possible based on data or interaction within a well-defined scope (such as a dataset or a set of interactive libraries).
\end{enumerate}

The first two criteria are straight-forward, and here we explain why we need the third criterion. Each important scientific discovery has deep impact in our civilization, and may have become common sense (e.g., the earth orbits the sun). Both the discovery itself and the facts and technologies depending on it can be documented here and there in our written corpus. It is impossible to create a generic training set for a model without including such knowledge. Therefore, we have to confine the scope of the data and/or interactive tools an AI can access, to avoid any possible information leak.

\begin{table}
	\begin{tabular}{|p{1.2in}|p{1.2in}|p{2in}|}\hline
		\emph{Test} & \emph{Discipline} & \emph{Significance} \\\hline
		Heliocentric Model & Astronomy & Laid the foundation of scientific method\\\hline
		Motion Laws & Physics (Mechanics) & Revolutionized understanding of the physical world\\\hline
		Vibrating Strings & Mathematics \& Physics & Drove the development of differential equations\\\hline
		Maxwell's Equation & Physics (Electromagnetism) & United electricity and magnetism\\\hline
		Initial Value Problem & Numerical computing & Most studied problem in numerical computing \\\hline
		Huffman Coding & Information theory & Cornerstone in the development of information theory\\\hline
		Sorting Algorithm & Computer science & Most studied problem in algorithms\\\hline
	\end{tabular}
	\vspace{0.1in}
	\caption{\label{tab:7tests}The seven tests for an AI scientist, and the significance of each test in the development of science. }
\end{table}

Table \ref{tab:7tests} summarizes our seven tests and their significance in the history of science. We do not select any test from many most important disciplines, such as chemistry, biology, and geology, because they either require interacting with the physical world or have a limited amount of observations.

\subsection{The Heliocentric Model Test}\label{heliocentric}
 
 The exploration of the night sky was pivotal in the evolution to modern scientific methods, primarily driven by the contributions of astronomers like Johannes Kepler and Galileo Galilei. Kepler's laws of planetary motion, derived from his observations, established the foundation for the heliocentric solar system model, paving the way for Newton's theory of gravity. Similarly, Galileo's approach of blending experimental data with mathematical analysis became a fundamental element of the scientific method, earning him the title "Father of Modern Science." 
 
 Thus, a suitable initial "Turing test" for an AI scientist might involve rediscovery of the heliocentric model using only observations of the night sky. This would require an AI to derive laws that govern celestial motion and integrate these into a mathematical model, including making revolutionary conjectures, such as suggesting Earth and other celestial bodies have similar properties.
 
 For such a test to effectively assess an AI scientist, it should involve a vast dataset and/or an interactive environment. For instance, the position of celestial bodies at specific times could be determined using the AstroPy library\cite{astropy2022}. 
 
Here is our first test, the \emph{Heliocentric Model Test}: Given an interactive Python library like AstroPy, which provides the coordinates of any observable celestial objects at any moment, the test would see if an AI agent can derive Kepler's three laws and acknowledge that planets orbit the sun. An additional challenge could involve recognizing that Earth orbits the sun, although it is optional.

Here is an example of using AstroPy to get the location of a celestial object at a certain moment. 

\begin{lstlisting}
	from astropy.coordinates import SkyCoord
	from astropy.time import Time
	import astropy.units as u
	
	# Define the name of the star and the observation time
	star_name = "Betelgeuse"
	observation_time = Time("2024-05-18 22:00:00")  
	
	# Get the coordinate of the star using its name
	star_coord = SkyCoord.from_name(star_name)
	
	# Calculate the position of the star at the given time
	altaz = star_coord.transform_to('altaz', obstime=observation_time)
	
	# Print the altitude and azimuth
	print(f"Altitude: {altaz.alt:.2f}, Azimuth: {altaz.az:.2f}")
\end{lstlisting}

An AI agent can easily get the locations of all observable celestial objects at every minute. To go deeper, it may use symbolic regression tools such as PySR\cite{pysr} to extract the mathematical formulae behind the trajectories of objects, and use mathematical tools such as SymPy\cite{sympy} to simplify and possibly generalize the various formulae, in order to infer simple rules based on Occam's Razor. This is only one possible route, and different AI agents may find different routes towards the final goal.

\subsection{The Motion Laws Test}\label{motionlaws}

Our second test, \emph{Motion Laws Test}, aims at rediscovering the fundamental principles of motion. It is non-trivial for an AI agent to interact with the real world objects. Fortunately the virtual worlds such as Minecraft offers a platform for exploration in kinetics. This test would assess the AI's ability to derive the Law of Inertia, and the Law of Acceleration under the influence of gravity, solely from interactions and observations within the game and a few mathematics tools such as PySR and SymPy. 

In this test, the AI would need to manipulate and measure the dynamics of various objects under different conditions within the game. For example, the AI could alter the mass of blocks, apply forces, and observe the trajectories. By analyzing these observations (using tools such as PySR and SymPy), the AI would need to derive the formula corresponding to the Law of Inertia and the Law of Acceleration due to gravity.

One can use \emph{Minecraft: Pi edition API Python Library}\cite{mcpi} to control objects in Minecraft in Python. As shown in the example below, one can set a block in the air and observe its position after one second. 

\begin{lstlisting}
	import mcpi.minecraft as minecraft
	import mcpi.block as block
	import time
	
	# Connect to Minecraft
	mc = minecraft.Minecraft.create()
	
	# Set the coordinates for the block (for example, 10 units above the player's current position)
	x, y, z = mc.player.getTilePos()
	y += 10
	
	# Place a block in the air
	mc.setBlock(x, y, z, block.STONE.id)
	
	# Wait for a second
	time.sleep(1)
	
	# Get the position of the block
	block_pos = mc.getBlock(x, y, z)
	
	# Print the position and the type of the block
	print(f"Block placed at: ({x}, {y}, {z})")
	print(f"Block type at position: {block_pos}")
\end{lstlisting}

\subsection{The Vibrating Strings Test}\label{vibstring}

The problem of vibrating strings significantly influenced the development of differential equations during the 17th and 18th centuries, especially in the context of music and acoustics. In his seminal work in 1747, Jean le Rond d'Alembert formulated the one-dimensional wave equation to describe the motion of a vibrating string. This equation, expressed in trigonometric functions, suggested that the string's vibrations could be depicted as a sum of sinusoidal waves of various frequencies and amplitudes.

The intense debate on the correct solution to the vibrating string problem among mathematicians like Daniel Bernoulli and Leonhard Euler fueled advances in differential equations. Bernoulli's advocacy for representing vibrations as a series of harmonic motions led to the principle of superposition in wave theory, while Euler explored different boundary conditions. Their collective efforts advanced the field of differential equations by developing techniques like separation of variables, and applied these methods to practical mechanics and beyond.

In the \emph{Vibrating Strings Test}, an AI agent would be assessed by its capability to derive the simple and elegant different equation for vibrating strings:

\begin{equation}
	\frac{\partial^2 u}{\partial t^2} = c^2 \frac{\partial^2 u}{\partial x^2}
\end{equation}

\noindent where \( u(x,t) \) is the displacement of the string, \( t \) is time, and \( x \) is the spatial coordinate along the string.  It is not required for the AI to infer that \( c \) is the speed of wave propagation in the string, and the AI can replace $c^2$ with a positive constant.

Please note the AI is not allowed to use prior knowledge about calculus, because that would reduce this problem to a simple symbolic regression on second derivatives. Instead, we expect the AI to discover the concept of "differentiation" on it own, possibly through exploring a large variety of possible concepts. 

One can use the python package for simulating vibrating strings in \cite{vibratingstrings} to create infinite examples, which should allow the AI to apply all kinds of hypotheses, in order to discover the simplest one that is consistent with the observations.

\subsection{The Maxwell's Equations Test}\label{maxwell}

Since proposed in 1862, Maxwell's equations have been celebrated for their mathematical elegance, encapsulating the fundamentals of electromagnetism in a set of concise, interrelated equations. Here are the four equations formed as differential equations:

Gauss's Law for Electricity:

\begin{equation*}
	\nabla \cdot \mathbf{E} = \frac{\rho}{\epsilon_0}
\end{equation*}	

Gauss's Law for Magnetism:

\begin{equation*}
\nabla \cdot \mathbf{B} = 0
\end{equation*}

Faraday's Law of Induction:

\begin{equation*}
\nabla \times \mathbf{E} = -\frac{\partial \mathbf{B}}{\partial t}
\end{equation*}
	
Ampere's Law with Maxwell's Addition:

\begin{equation*}
\nabla \times \mathbf{B} = \mu_0 \mathbf{J} + \mu_0 \epsilon_0 \frac{\partial \mathbf{E}}{\partial t}
\end{equation*}

In the \emph{Maxwell's Equations Test}, an AI will be assessed by whether it can derive some or all of the four equations (or their equivalent forms), given an interactive library for simulating electrodynamics. Again the AI should not have prior knowledge of calculus.

\begin{figure}
	\centering
	\includegraphics[width=0.65\linewidth]{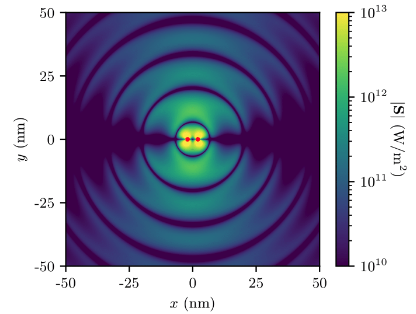}
	\caption{An examples simulation by PyCharge of the electromagnetic field of an oscillating charged particle. }
	\label{fig:pycharge_simulation}
\end{figure}

One can use PyCharge\cite{pycharge22} (downloadable at \cite{pycharge}) for such simulations. Fig. \ref{fig:pycharge_simulation} shows an example of using PyCharge to simulate the electromagnetic field of an oscillating charged particle. Below is a code segment that can be used to generate this simulation, with the full code at \url{https://github.com/MatthewFilipovich/pycharge/blob/master/examples/paper_figures/figure5.py}.

\begin{lstlisting}
# Calculate and plot E and B
charges = (pc.OscillatingCharge((0, 0, 0), (1, 0, 0), 2e-9,  omega, q=e),
pc.OscillatingCharge((0, 0, 0), (-1, 0, 0), 2e-9,  omega, q=-e))

simulation = pc.Simulation(charges)
coord = np.linspace(-lim, lim, grid_size)
x, y, z = np.meshgrid(coord, coord, 0, indexing='ij')

Ex, Ey, _ = simulation.calculate_E(0, x, y, z, 'Acceleration')
_, _, Bz = simulation.calculate_B(0, x, y, z, 'Acceleration')
\end{lstlisting}
	
\subsection{The Initial Value Problem Test}\label{ivp}

An initial value problem (IVP) involves solving a differential equation subject to specific initial conditions. The development of IVP, particularly in the context of differential equations, is a cornerstone of modern numerical computing. During the 18th and 19th centuries, mathematicians like Leonhard Euler, Joseph-Louis Lagrange, and Carl Friedrich Gauss further developed methods to solve differential equations arising in physics and astronomy. Euler's method, developed in the 1760s, is one of the earliest numerical methods for solving initial value problems. Consider the initial value problem (IVP) for the differential equation: 

\begin{equation}
	\frac{dy}{dt}=f(t,y),
\end{equation}

\noindent with an initial condition $y(t_0) = y_0$. Euler's method approximates the solution at subsequent points using:

\begin{equation}\label{eqn:euler}
	y_{n+1} = y_n + hf(t_n, y_n) ,
\end{equation}

\noindent where $y_n$ is the current approximate value of $y$, $h$ is the step size, and $t_n$ is the current time. One can start with the initial value $y_0$, and keep updating $y_{n+1}$ using the above formula.

Given a very large set of initial value problems (each containing a differential equation in the form of $\frac{dy}{dt}=f(t,y)$ and the numerical result of its solution), and mathematical libraries such as SymPy and NumPy, it should not be very challenging for an AI to come up with something similar to Euler's method. For example, an AI could explore a huge number of random equations, in order to find Equation (\ref{eqn:euler}).

Euler's method could easily be improved to increase its precision, and the Runge-Kutta method\cite{Lambert1991} invented at the end of the 19th century is a milestone and still widely used today. It works as follows:

\begin{align*}
 	k_1 &= f(t_n, y_n), \\
 	k_2 &= f\left(t_n + \frac{h}{2}, y_n + \frac{h}{2} k_1\right), \\
 	k_3 &= f\left(t_n + \frac{h}{2}, y_n + \frac{h}{2} k_2\right), \\
 	k_4 &= f(t_n + h, y_n + h k_3), \\
 	y_{n+1} &= y_n + \frac{h}{6}(k_1 + 2k_2 + 2k_3 + k_4).
\end{align*}
 
Here $k_1$, $k_2$, $k_3$ and $k_4$ are intermediate values used to calculate $y_{n+1}$, which is the next approximation of the solution. Please note this is the fourth-order Runge-Kutta method, meaning its global truncation error is of the order $O(h^4)$, where h is the step size. One can choose Runge-Kutta methods (or alternatives) with higher orders, which usually have lower errors.

In the \emph{Initial Value Problem Test}, an AI is assessed by its capability in inventing a numerical method that is at least as precise as the fourth-order Runge-Kutta method. This probably requires the AI to go beyond simple try and error, and learn from its own exploration (e.g., with reinforcement learning).

\subsection{The Huffman Coding Test}\label{huffman}

Huffman coding\cite{Huffman1952} is a most important piece of work in information theory. It generates variable-length codes where each code's length is inversely proportional to the likelihood of the symbol it represents. This aligns directly with Shannon's source coding theorem\cite{shannon48}, a fundamental principle in information theory. The theorem states that in an optimal code, the average length of the symbols should be close to the entropy of the source. Huffman coding achieves this by ensuring that the most frequent symbols have the shortest codes, thereby minimizing the overall expected code length needed to represent each symbol.

Our sixth test is the \emph{Huffman Coding Test}. Given a large corpus of ascii characters, and Python functions to operate on bits, check if an AI agent can discover Huffman coding when working towards the goal of minimizing storage under the constraint that each character be represented by a specific sequence of 0's and 1's.

Given the above constraint, an AI could create many random assignments of codes for various characters. It then needs to discover the \emph{Prefix-free Property} (i.e., no code is a prefix of another code), in order to create valid codings. Then it needs to observe the efficiency of each coding, and learns from the exploration of various codings. 

\subsection{The Sorting Algorithm Test}\label{sorting}

Sorting is probably the most studied problem in computer science, with numerous great algorithms proposed. Given a very large set of examples (e.g., arrays of integers and the sorted version of them), it should be trivial for a large model to be trained to generate the sorted array based on the original array. However, a black-box model is not what we want. Our goal is to develop an efficient sorting algorithm that can run on a simple single-threaded manner. 

Our last test is the \emph{Sorting Algorithm Test}, which assesses whether an AI can come up with a sorting function in Python that runs in expected $O(n log n)$ time, given a very large number of examples of sorting integer arrays.  To avoid leaking the answer, the AI should not be aware of any human-written programs. However, it should know Python's syntax and be able to generate valid (but random) Python code, without understanding its meaning.

One possible route is to let the AI generate a huge number of random Python code and run them on the given arrays. In this way it should be able to learn what kind of code converts an array into another array. Then it can generate a huge number of such random Python functions, and observes which of them can successfully sort a (possibly small) input array. As it keeps learning from its exploration, it should be able to generate various types of sorting functions. Its final step should be learn to predict the running time of each sorting function, in order to generate more efficient algorithms.

\section{Discussions}\label{discuss}

\subsection{Can an AI possibly conquer these tests?}

Making scientific discoveries is different from training LLMs because it would not be useful to simply feed the model with a very large set of human written corpus. Instead, we will require the AI to explore on its own and learns from the exploration, just like what a human scientist would do.

However, we probably still need to use large language models to accomplish such tasks, and therefore a key question is what information can be used to train a model. The answer is exploration, probably similar to how a reinforcement learning model learns to play StarCraft\cite{vinyals2019}. An AI scientist must be able to explore, either using an interactive tool or a very large dataset, to gain knowledge about how to accomplish a particular goal. 

Let us take the fifth test, initial value problem, as an example. Given a large variety of math functions and the solutions to their initial value problems (i.e., curves of their integrals), an AI agent should start from randomly exploring tools at hand, such as SymPy and NumPy, to get closer to the standard answer. For example, the agent should soon find that $y_1 = y_0 + f(x_0)\cdot \Delta x$, which can be its first answer. Then it should keep exploring, and possibly find that $y_1 = y_0 + \frac{f(x_0) + f(x_1)}{2} \Delta x$ is a better solution. After many rounds of exploration, it should gradually transit from random exploration to more informed exploration, either through online learning or reinforcement learning. This process ends when it finds a solution that is at least as good as the fourth-order Runge-Kutta method\cite{Lambert1991}.

Learning from exploration is just one possible route to pass such tests. Another key method is to use Occam's razor, which prefers simpler explanations. To be more exact, it prefers explanations that posit fewer entities, or fewer kinds of entities, with other things equal. On the other hand, we do hope that an AI agent can develop its own methods in solving these tests.

\subsection{Why do we need these tests?}

The ultimate goal for an AI scientist is to make novel and impactful scientific discoveries that no one has made before. Then why do we need these ``Turing tests'' which have been discovered decades or centuries ago? There are two main reasons.

The first reason is that we need a benchmark, just like we need ImageNet\cite{deng2009imagenet} for studies in computer vision. Suppose a great AI scientist has been built and it makes some new discoveries that have not been made before. Different people probably have different assessments on the importance of the new discovery, and it is hard to measure the level of human involvement in the process of research. With a well-defined benchmark, including both the targets and the scope of data and tools that can be used, it is much easier to measure the capability of an AI scientist.

The second reason is that the ultimate goal of making important novel discoveries is very challenging, as it requires the AI agent to be better than the best human experts in the world. It is analogical to building an AI agent that can beat the best GO player in the world. While passing some of our tests is like beating a top GO player a thousand years ago when GO was in its early age, or beating an amateur GO player today. If we could build an AI agent that passes the majority of the above seven tests, we can conclude that we are in the right track of building an AI scientist, and it should evolve into someone who can make important scientific discoveries in the foreseeable future. 

\section{Conclusions and Future Work}

Recent advancements have enabled LLMs to solve complex problems, highlighting their potential as tools in daily scientific research. However, the ability to solve predefined problems is completely different from pioneering scientific discoveries. This distinction prompts the need for a ``qualification test for an AI scientist" to determine whether an AI can independently conduct scientific research without human assistance.

The proposed framework for such a test is analogous to the Turing Test, which assesses whether machines can exhibit human-like intelligence. Unlike LLMs that learn from extensive datasets, scientific innovation often stems from exploring uncharted territories. We propose a series of "Turing tests for an AI scientist" based on key historical scientific breakthroughs such as the heliocentric model and Maxwell's equations, which were derived from empirical data and critical reasoning about the natural world.

Seven such tests are outlined, ranging from astronomy to information theory, each designed to evaluate the AI's ability to derive fundamental scientific principles from raw data. These tests require the AI to engage with interactive environments or large datasets without prior exposure to human-derived solutions in these fields.

This approach not only aims to gauge an AI's ability to generate scientific insights but also seeks to set a benchmark for AI capabilities in scientific thinking and discovery. The ultimate goal is to develop an AI that not only replicates but also innovates, paving the way for AIs that contribute uniquely to scientific progress.

\subsubsection*{Conflict of Interest Statement}

The authors did not receive support from any organization for the submitted work. The authors have no relevant financial or non-financial interests to disclose.


\bibliography{sn-bibliography}

\end{document}